\documentclass[11pt,a4paper]{article}
\usepackage[hyperref]{naaclhlt2018}
\usepackage{times}
\usepackage{latexsym}
\usepackage{graphicx}
\usepackage{amsmath}
\usepackage{amssymb}
\usepackage{color}
\usepackage{booktabs}
\usepackage{xspace}

\usepackage{url}

\def\figref#1{Fig.~\ref{#1}}
\def\secref#1{Sec.~\ref{#1}}
\def\tabref#1{Table~\ref{#1}}

\aclfinalcopy 
\def\aclpaperid{1074} 

\newcommand\AddSentTrain[0]{\mbox{AddSentDiverse}\xspace}

\title{Robust Machine Comprehension Models via Adversarial Training}

\author{
	Yicheng Wang \and Mohit Bansal \\
    University of North Carolina at Chapel Hill \\
    {\tt \{yicheng, mbansal\}@cs.unc.edu}
}

\date{}

\begin{document}
\maketitle

\begin{abstract}
It is shown that many published models for the Stanford Question Answering Dataset~\cite{SQuAD} lack robustness, suffering an over 50\% decrease in F1 score during adversarial evaluation based on the AddSent~\cite{advSQuAD} algorithm. It has also been shown that retraining models on data generated by AddSent has limited effect on their robustness. We propose a novel alternative adversary-generation algorithm, \AddSentTrain, that significantly increases the variance within the adversarial training data by providing effective examples that punish the model for making certain superficial assumptions. Further, in order to improve robustness to AddSent's semantic perturbations (e.g., antonyms), we jointly improve the model's semantic-relationship learning capabilities in addition to our \AddSentTrain-based adversarial training data augmentation. With these additions, we show that we can make a state-of-the-art model significantly more robust, achieving a 36.5\% increase in F1 score under many different types of adversarial evaluation while maintaining performance on the regular SQuAD task.
\end{abstract}

\section{Introduction}

We explore the task of reading comprehension based question answering (Q\&A), where we focus on the Stanford Question Answering Dataset (SQuAD) \cite{SQuAD}, in which models answer questions about paragraphs taken from Wikipedia. Significant progress has been made with deep end to end neural-attention models, with some achieving above human level performance on the test set~\cite{mLSTM,BiDAF,rnet,Huang17,ELMo}. However, as shown recently by~\citet{advSQuAD}, these models are very fragile when presented with adversarially generated data. They proposed AddSent, which creates a semantically-irrelevant sentence containing a fake answer that resembles the question syntactically, and appends it to the context. Many state-of-the-art models exhibit a nearly 50\% reduction in F1 score on AddSent, showing their over-reliance on syntactic similarity and limited semantic understanding.

Importantly, this is in part due to the nature of the SQuAD dataset. Most questions in the dataset have answer spans embedded in sentences that are syntactically similar to the question. Thus during training, the model is rarely punished for answering questions based on syntactic similarity, and learns it as a reliable approach to Q\&A. This correlation between syntactic similarity and correctness is of course not true in general: the adversaries generated by AddSent~\cite{advSQuAD} are syntactically similar to the question but do not answer them. The models' failures on AddSent demonstrates their ignorance of this aspect of the task. \citet{advSQuAD} presented some initial attempts to fix this problem by retraining the BiDAF model \cite{BiDAF} with adversaries generated with AddSent. But they showed that the method is not very effective, as slight modifications (e.g., different positioning of the distractor sentence in the paragraph and different fake answer set) to the adversary generation algorithm at test time have drastic impact on the retrained model's performance.

In this paper, we show that their method of adversarial training failed because the specificity of the AddSent algorithm along with the lack of naturally-occurring counterexamples allow models to learn superficial clues regarding what is a `distractor' and subsequently ignore it; thus significantly limiting their robustness. Instead, we first introduce a novel algorithm, \AddSentTrain, for generating adversarial examples with significantly higher variance (by varying the locations where the distractors are placed and expanding the set of fake answers), so that the model is punished during training time for making these superficial assumptions about the distractor. We show that an \AddSentTrain-based adversarially-trained model beats an AddSent-trained model across 3 different adversarial test sets, showing an average improvement of 24.22\% in F1 score, demonstrating a general increase in robustness.

However, even with our diversified adversarial training data, the model is still not fully resilient to AddSent-style attacks, e.g., its antonymy-style semantic perturbations. Hence, we next add semantic relationship features to the model to let it directly identify such relationships between the context and question. Interestingly, we see that these additions only increase model robustness when trained adversarially, because intuitively in the non-adversarially-trained setup, there are not enough negative (adversarial) examples for the model to learn how to use its semantic features.

Overall, we demonstrate that with our adversarial training method and model improvement, we can increase the performance of a state-of-the-art model by 36.46\% on the AddSent evaluation set. Although we focused on the AddSent adversary~\cite{advSQuAD}, our method of effective adversarial training by eliminating superficial statistical correlations (with joint model capability improvements) are generalizable to other similar insertion-based adversaries for Q\&A tasks.\footnote{We release our \AddSentTrain-based adversarial training dataset for SQuAD at \url{https://goo.gl/qdSNDr}.}
\section{Related Work}

\paragraph{Adversarial Evaluation}

In computer vision, adversarial examples are frequently used to punish model oversensitivity, where semantic-preserving perturbations (usually in the form of small noise vectors) are added to an image to fool the classifier into giving it a different label~\cite{advImg, advImgTrain}.

In the field of Q\&A, \citet{advSQuAD} introduced the AddSent algorithm, which generates adversaries that punish model failure in the other direction: overstability, or the inability to detect semantic-altering noise. It does so by generating distractor sentences that only resemble the questions syntactically and appending them to the context paragraphs (detailed description included in \secref{sec:method}). When tested on these adversarial examples, \citet{advSQuAD} showed that even the most `robust' amongst published models (the Mnemonic Reader \cite{mmr}) only achieved 46.6\% F1 (compared to 79.6\% F1 on the regular task). Since then, the FusionNet model \cite{Huang17} used history-of-word representations and multi-level attention mechanism to obtain an improved 51.4\% F1 score under adversarial evaluation, but that is still a 30\% decrease from the model's performance on the regular task. We show, however, that one can make a pre-existing model significantly more robust by simply retraining it with better, higher variance adversarial training data, and improve it further with minor semantic feature additions to its inputs.

\paragraph{Adversarial Training}

It has been shown in the field of image classification that training with adversarial examples produces more robust and error-resistant models~\cite{advImgTrain,advImgNet}. In the field of Q\&A,~\citet{advSQuAD} attempted to retrain the BiDAF \cite{BiDAF} model with data generated with AddSent algorithm. Despite performing well when evaluated on AddSent, the retrained model suffers a more than 30\% decrease in F1 performance when tested on a slightly different adversarial dataset generated by AddSentMod (which differs from AddSent in two superficial ways: using a different set of fake answers and prepending instead of appending the distractor sentence to the context). We show that using AddSent to generate adversarial training data introduces new superficial trends for a model to exploit; and instead we propose the \AddSentTrain algorithm that generates highly varied data for adversarial training, resulting in more robust models.
\section{Methods}
\label{sec:method}

Our `\AddSentTrain' algorithm is a modified version of AddSent \cite{advSQuAD}, aimed at producing good adversarial examples for robust training purposes. For each \{context, question, answer\} triple, AddSent does the following: (1) Several antonym and named-entity based semantic altering perturbations (swapping) are applied to the question; (2) A fake answer is generated that matches the `type' of the original answer (e.g., Prague $\to$ Chicago, etc.); (3) The fake answer and the altered question are combined into a distractor statement based on a set of manually defined rules; (4) Errors in grammar are fixed by crowd-workers; (5) The finalized distractor is appended to the end of the context. The specificity of the algorithm creates new superficial cues that a model can learn and use during training and never get punished for: (1) a model can learn that it is unlikely for the last sentence to contain the real answer; (2) a model can learn that the fixed set of fake answers should not be picked. These nullify the effectiveness of the distractors as the model will learn to simply ignore them. We thus introduce the \AddSentTrain algorithm, which adds two modifications to AddSent that allows for generating higher-variance adversarial examples. Namely, we randomize the distractor placement (\secref{sec:dist_placement}) and we diversity the set of fake answers used (\secref{sec:fake_answers}). Lastly, to address the antonym-style semantic perturbations used in AddSent, we show that we need to improve model capabilities by adding indicator features for semantic relationships (but only when) in tandem with the addition of diverse adversarial data (\secref{sec:feature_models}).

\subsection{Random Distractor Placement}
\label{sec:dist_placement}
\begin{figure}[t]
	\centering
    \includegraphics[clip,trim=3.5cm 0 0 0,width=8.5cm]{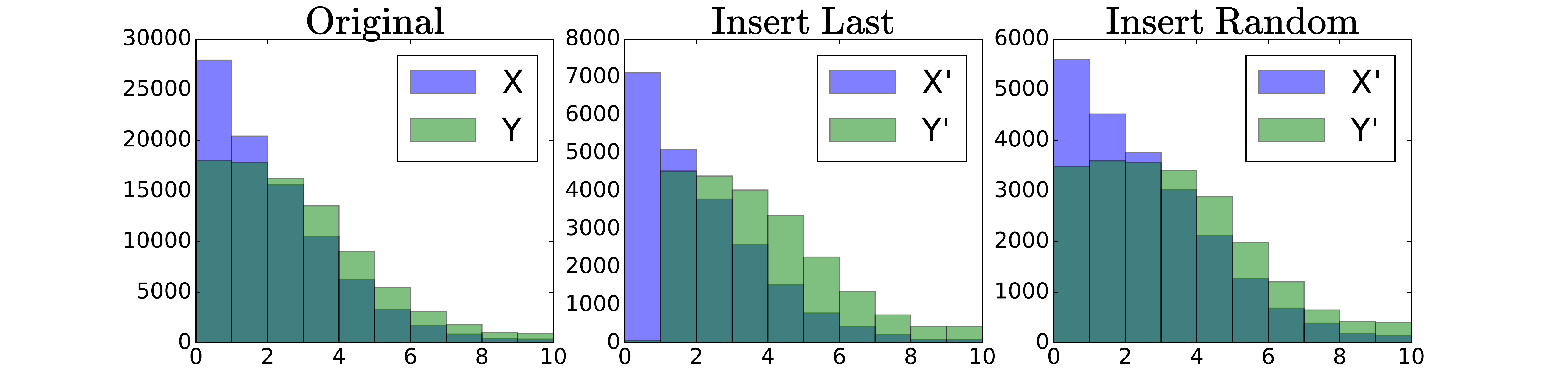}
    \vspace{-20pt}
    \caption{Left: Distribution of $X$ and $Y$ for the original SQuAD training set.
    Middle: Distribution of $X'$ and $Y'$ when the distractor is inserted
    at the end of the context.
    Right: Distribution of $X'$ and $Y'$ when the distractor is inserted
    randomly into the context.}\label{fig:plot1}\label{fig:plot2}
\end{figure}

Given a paragraph $P$ containing $n$ sentences, let $X$, $Y$ be random variables representing the location of the sentence containing the correct answer counting from the front and back.\footnote{Note that for any fixed $n$, $Y = n - X$, but for our purposes it is easier to keep them separate since the length of the paragraph is also a random variable.} Let $P'$ represent the paragraph with the inserted distractor, and $X'$ and $Y'$ represent the updated location of the sentence with the correct answer. As shown in \figref{fig:plot1}, their distribution is highly dependent on the strategy used to insert the distractor. During training done by \citet{advSQuAD}, the distractor is always added as the last sentence, creating a very skewed distribution for $Y'$. This resulted in the model learning to ignore the last sentence, as it was never punished for doing so. This, in turn, caused the retrained model to fail on AddSentMod, where the distractor is inserted to the front instead of the back of the context paragraph (this is shown by our experiments as well). However, \figref{fig:plot1} shows that when the distractor is inserted randomly, the distributions of $X'$ and $Y'$ are almost identical to that of $X$ and $Y$, indicating that no new correlation between the location of a sentence and its likelihood to contain the correct answer is introduced by the distractors, hence forcing the model to learn to discern them from the real answers by other, deeper means.

\begin{table*}[t]
    \centering
    \begin{small}
    \begin{tabular}{c|ccccc|c}
        Training & Original-SQuAD-Dev & AddSent &  AddSentPrepend & AddSentRandom &
        AddSentMod & Average \\\hline
        Original-SQuAD & \textbf{84.65} & 42.45 & 41.46 & 40.48 & 41.96 & 50.20 \\
        AddSent & 83.76 & \textbf{79.55} & 51.96 & 59.03 & 46.85 &  64.23 \\
        \AddSentTrain & 83.49 & 76.95 & \textbf{77.45} & \textbf{76.02} & \textbf{77.06} & \textbf{78.19}
    \end{tabular}
    \vspace{-12pt}
    \caption{F1 performance of the BSAE model trained and tested on different regular/adversarial datasets.\vspace{-10pt}}\label{tab:table1}
        \end{small}
\end{table*}

\begin{table}[t]
    \centering
    \begin{small}
    \begin{tabular}{c|ccc}
        Training & AddSent & AddSentPrepend & Average \\
        \hline
        InsFirst & 60.22 & \textbf{79.81} & 70.02 \\
        InsLast & \textbf{79.54} & 51.96 & 	65.75 \\
        InsMid & 74.74 & 74.33 & 74.54 \\
        InsRandom & 76.33 & 77.38 & \textbf{76.85} \\
    \end{tabular}
    \vspace{-7pt}
    \caption{F1 performance of the BSAE model trained on datasets with different
    distractor placement strategies.}
    \label{tab:table2}
    \end{small}
\end{table}

\subsection{Dynamic Fake Answer Generation}
\label{sec:fake_answers}

To prevent the model from superficially deciding what is a distractor based on certain specific words, we dynamically generate the fake answers instead of using AddSent's pre-defined set. Let $S$ be the set that contains all the answers in the SQuAD training data, tagged by their type (e.g., person, location, etc.). For each answer $a$, we generate the fake answer dynamically by randomly selecting another answer $a' \neq a$ from $S$ that has the same type as $a$, as opposed to AddSent~\cite{advSQuAD}, which uses a pre-defined fake answer for each type (e.g., ``Chicago'' for any location). This creates a much larger set of fake answers, thus decreasing the correlation between any text and its likelihood of being a part of a distractor, forcing the model to become more robust.

\subsection{Semantic Feature Enhanced Model}
\label{sec:feature_models}

In previous sections, we prevented the model from identifying distractors based on superficial clues such as location and fake answer identity by eliminating these correlations within the training data. But even if we force the model to learn some deeper methods for identifying/discarding the distractors, it only has limited ability in recognizing semantic differences because its current inputs do not capture crucial aspects of lexical semantics such as antonymy (which were inserted by~\newcite{advSQuAD} when generating the AddSent adversaries; see Sec.~\ref{sec:method}). Most current models use pretrained word embeddings (e.g., GloVE~\cite{GloVE} and ELMo~\cite{ELMo}) as input, which are usually calculated based on the distributional hypothesis~\cite{dist_hypo}, and do not capture lexical semantic relations such as antonymy~\cite{ido_dagan_dist_lexical_entail}. These shortcomings are reflected by our results in \secref{sec:ant_model_results}, where we see that we can't resolve all AddSent-style adversaries by diversifying the training data alone. For the model to be robust to semantics-based (e.g., antonym-style) attacks, it needs extra knowledge of lexical semantic relations. Hence, we augment the input of each word in the question/context with two indicator features indicating the existence of its synonym and antonym (using WordNet~\cite{WordNet}) in the context/question, allowing the model to use lexical semantics directly instead of learned statistical correlations of the word embeddings.
\section{Experiments And Results}
\label{sec:experiments}

\subsection{Model and Training Details}

We use the architecture and hyperparameters of the strong BiDAF + Self-Attn + ELMo (BSAE) model~\cite{ELMo}, currently (as of January 10, 2018) the third highest performing single-model on the SQuAD leaderboard.\footnote{{\tiny\url{https://rajpurkar.github.io/SQuAD-explorer/}}}

\subsection{Evaluation Details}
Models are evaluated on the original SQuAD dev set and 4 adversarial datasets: AddSent, the adversarial evaluation set by \citet{advSQuAD}, and 3 variations of AddSent: AddSentPrepend, where the distractor is prepended to the context, AddSentRandom, where the distractor is randomly inserted into the context,\footnote{Note that since the distractor was randomly inserted, the model cannot identify/ignore the distractor reliably based on location. Thus, high performance on AddSentRandom serves as a better indicator for robustness to semantic-based attacks.} and AddSentMod \cite{advSQuAD}, where a different set of fake answers is used and the distractor is prepended to the context. Experiments measure the soft F1 score and all of the adversarial evaluations are model-dependent, following the style of AddSent, where multiple adversaries are generated for each example in the evaluation set and the model's worst performance among the variants is recorded.

\subsection{Primary Experiment Results}

In our main experiment, we compare the BSAE model's performance on different test sets when trained with three different training sets: the original SQuAD data (Original-SQuAD), SQuAD data augmented with AddSent generated adversaries (similar to adversarial training conducted by \citet{advSQuAD}), and SQuAD data augmented with our \AddSentTrain generated adversaries. For the latter two, we run the respective adversarial generation algorithms on the training set, and add randomly selected adversarial examples such that they make up 20\% of the total training data. The results are shown in \tabref{tab:table1}. First, as shown, the AddSent-trained model is not able to perform well on test sets where the distractors are not inserted at the end, e.g., the AddSentRandom adversarial test set.\footnote{For this 59.03\% accuracy, i.e., in the remaining 40.96\% errors, we found that in 77.0\% of these errors, the model still predicted a span within the randomly inserted distractor; indicating that it has not learned to fully recognize semantic-altering perturbations.} On the other hand, it can be seen that retraining with \AddSentTrain boosts performance of the model significantly across all adversarial datasets, indicating a general increase in robustness.

\subsection{Distractor Placement Results}

We also conducted experiments studying the effect of different distractor placement strategies on the trained models' robustness. The BSAE model was trained on 4 variations of \AddSentTrain-augmented training set, with the only difference between them being the location of the distractor within the context: InsFirst, where the distractor is prepended, InsLast, where the distractor is appended, InsMid, where the distractor is inserted in the middle and InsRandom, where the distractor is randomly placed. The retrained models are tested on AddSent and AddSentPrepend, whose only difference is where the distractor is located. The result is shown in \tabref{tab:table2}. It is clear that when trained under InsFirst and InsLast, the model only performs well on test sets created by a similar distractor placement strategy, indicating that they are exploiting superficial trends instead of learning to process the semantics. It is also shown that InsRandom gives optimal performance on both evaluation datasets. Further investigations regarding distractor placement can be found in the appendix.

\begin{table}[t]
    \centering
    \begin{small}
    \begin{tabular}{c|cc}
        Training & AddSentPrepend & AddSentMod\\\hline
        Fixed-FakeAns & 77.37 & 73.65 \\
        Dynamic-FakeAns & \textbf{77.45} & \textbf{77.06}
    \end{tabular}
    \vspace{-5pt}
    \caption{F1 performance of the BSAE model trained on datasets with different
    answer generation strategies.}\label{tab:table3}
    \vspace{-2pt}
    \end{small}
\end{table}

\subsection{Fake Answer Generation Results}

We also conducted experiments studying the effect of training on data containing distractors with dynamically generated fake answers (Dynamic-FakeAns) instead of chosen from a predefined set (Fixed-FakeAns). The trained models are tested on AddSentPrepend and AddSentMod, whose only difference is that AddSentMod uses a different set of fake answers. The results are displayed in \tabref{tab:table3}. It shows that the model trained on Fixed-FakeAns suffers an approximate 3\% drop in performance when tested on a dataset with a different set of fake answers, but this gap does not exist for the model retrained on Dynamic-FakeAns.

\begin{table}[t]
	\centering
    \begin{small}
    \begin{tabular}{c|c|c}
    Model/Training & Original-SQuAD-Dev & AddSent \\\hline
    BSAE/Reg. & \textbf{84.65} & 42.45 \\
    BSAE/Adv. & 83.49 & 76.95 \\
    BSAE+SA/Reg. & 84.62 & 44.60 \\
    BSAE+SA/Adv. & 84.49 & \textbf{78.91} \\
    \end{tabular}
    \caption{Regular and adversarial training with BSAE and BSAE+SA (with synonym/antonym features).}
    \label{tab:imp_of_features_and_training}
    \vspace{-7pt}
    \end{small}
\end{table}

\subsection{Semantic Feature Enhancement Results}
\label{sec:ant_model_results}

In \tabref{tab:table1}, we see that despite improving performance on adversarial test sets, adversarial training on the BSAE model leads to a 1\% decrease in its performance on the original SQuAD task (from 84.65\% to 83.49\%). Furthermore, there is still a 6.5\% gap between its performance on adversarial datasets and the original SQuAD dev set (76.95\% vs 83.49\%). These point to the limitations of adversarial training without any model enhancements, especially for AddSent's antonymy style semantic perturbations (see details in \secref{sec:feature_models}). We thus conducted experiments to test the effectiveness of adding WordNet based synonymy/antonymy semantic-relation indicators in helping the model to better deal with semantics-based adversaries. We added the lexical semantic indicators to the BSAE model to create the BSAE+SA model. We trained and tested it in both the regular and adversarial setup. Its results, compared to the original BSAE model are shown in \tabref{tab:imp_of_features_and_training}, where we see that unlike the BSAE model, adversarial training of the BSAE+SA model does not cause a decrease in its performance on the original SQuAD dataset, as the model can now learn lexical semantic relationships instead of statistical correlations. We also see that the BSAE+SA model, when trained in the normal setup, shows very similar performance as the BSAE model across all metrics. This is most likely because despite having the ability to recognize semantic relations, there are not enough negative examples in the regular SQuAD training set to teach the model how to use these features correctly, but this issue is solved via the addition of adversarial examples in adversarial training.

\subsection{Error Analysis}

Finally, we examined the errors of our final adversarially-trained BSAE+SA model on the AddSent dataset and found that out of the 21.09\% remaining errors (\tabref{tab:imp_of_features_and_training}), 33.3\% (46 cases) of these erroneous predictions occurred within the inserted distractor, and 63.7\% (88 cases) occurred on questions that the model got wrong in the original SQuAD dev set (without the inserted distractors). The former errors are mainly occurring within distractors created with named-entity replacements (which we haven't addressed directly in the current paper) or malformed distractors (that in fact do answer the question).
\section{Conclusion}

We demonstrate that we can overcome model overstability and increase their robustness by training on diverse adversarial data that eliminates latent data correlations. We further show that adversarial training is more effective when we jointly add useful semantic-relations knowledge to improve model capabilities. We hope that these robustness methods are generalizable to other insertion-based adversaries for Q\&A tasks.

\section*{Acknowledgments}
\vspace{-5pt}
We thank the anonymous reviewers for their helpful comments. This work was supported by a
Google Faculty Research Award, a Bloomberg Data Science Research Grant, an IBM Faculty
Award, and NVidia GPU awards.

\bibliography{mybib}

\begin{thebibliography}{}
\expandafter\ifx\csname natexlab\endcsname\relax\def\natexlab#1{#1}\fi

\bibitem[{Fellbaum(1998)}]{WordNet}
C.~Fellbaum. 1998.
\newblock Wordnet: An electronic lexical database.
\newblock In {\em MIT Press\/}.

\bibitem[{Geffet and Dagan(2005)}]{ido_dagan_dist_lexical_entail}
Maayan Geffet and Ido Dagan. 2005.
\newblock The distributional inclusion hypotheses and lexical entailment.
\newblock In {\em ACL\/}.

\bibitem[{Goodfellow et~al.(2015)Goodfellow, Shlens, and Szegedy}]{advImgTrain}
I.~Goodfellow, J.~Shlens, and C.~Szegedy. 2015.
\newblock Explaining and harnessing adversarial examples.
\newblock In {\em International Conference on Learning Representations
  (ICLR)\/}.

\bibitem[{Harris(1954)}]{dist_hypo}
Zellig~S Harris. 1954.
\newblock Distributional structure.
\newblock {\em Word\/} 10(2-3):146--162.

\bibitem[{Hu et~al.(2017)Hu, Peng, and Qiu}]{mmr}
Minghao Hu, Yuxing Peng, and Xipeng Qiu. 2017.
\newblock Reinforced mnemonic reader for machine comprehension.
\newblock {\em CoRR, abs/1705.02798\/} .

\bibitem[{Huang et~al.(2018)Huang, Zhu, Shen, and Chen}]{Huang17}
Hsin-Yuan Huang, Chenguang Zhu, Yelong Shen, and Weizhu Chen. 2018.
\newblock Fusionnet: Fusing via fully-aware attention with application to
  machine comprehension.
\newblock In {\em International Conference on Learning Representations
  (ICLR)\/}.

\bibitem[{Jia and Liang(2017)}]{advSQuAD}
R.~Jia and P.~Liang. 2017.
\newblock Adversarial examples for evaluating reading comprehension systems.
\newblock In {\em Empirical Methods in Natural Language Processing (EMNLP)\/}.

\bibitem[{Kurakin et~al.(2017)Kurakin, Goodfellow, and Bengio}]{advImgNet}
A.~Kurakin, I.~Goodfellow, and S.~Bengio. 2017.
\newblock Adversarial machine learning at scale.
\newblock In {\em International Conference on Learning Representations
  (ICLR)\/}.

\bibitem[{Pennington et~al.(2014)Pennington, Socher, and Manning}]{GloVE}
Jeffrey Pennington, Richard Socher, and Christopher~D. Manning. 2014.
\newblock Glove: Global vectors for word representation.
\newblock In {\em Empirical Methods in Natural Language Processing (EMNLP)\/}.
  pages 1532--1543.

\bibitem[{Peters et~al.(2018)Peters, Neumann, Iyyer, Gardner, Clark, Lee, and
  Zettlemoyer}]{ELMo}
Matthew~E Peters, Mark Neumann, Mohit Iyyer, Matt Gardner, Christopher Clark,
  Kenton Lee, and Luke Zettlemoyer. 2018.
\newblock Deep contextualized word representations.
\newblock {\em NAACL\/} .

\bibitem[{Rajpurkar et~al.(2016)Rajpurkar, Zhang, Lopyrev, and Liang}]{SQuAD}
P.~Rajpurkar, J.~Zhang, K.~Lopyrev, and P.~Liang. 2016.
\newblock Squad: 100,000+ questions for machine comprehension of text.
\newblock In {\em Empirical Methods in Natural Language Processing (EMNLP)
  2016\/}.

\bibitem[{Seo et~al.(2017)Seo, Kembhavi, Farhadi, and Hajishirzi}]{BiDAF}
Minjoon Seo, Aniruddha Kembhavi, Ali Farhadi, and Hannaneh Hajishirzi. 2017.
\newblock Bi-directional attention flow for machine comprehension.
\newblock In {\em International Conference on Learning Representations
  (ICLR)\/}.

\bibitem[{Szegedy et~al.(2014)Szegedy, Zaremba, Sutskever, Bruna, Erhan,
  Goodfellow, and Fergus}]{advImg}
C.~Szegedy, W.~Zaremba, I.~Sutskever, J.~Bruna, D.~Erhan, I.~Goodfellow, and
  R.~Fergus. 2014.
\newblock Intriguing properties of neural networks.
\newblock In {\em International Conference on Learning Representations
  (ICLR)\/}.

\bibitem[{Wang and Jiang(2017)}]{mLSTM}
S.~Wang and J.~Jiang. 2017.
\newblock Machine comprehension using match-lstm and answer pointer.
\newblock In {\em International Conference on Learning Representations
  (ICLR)\/}.

\bibitem[{Wang et~al.(2017)Wang, Yang, Wei, Chang, and Zhou}]{rnet}
Wenhui Wang, Nan Yang, Furu Wei, Baobao Chang, and Ming Zhou. 2017.
\newblock Gated self-matching networks for reading comprehension and question
  answering.
\newblock In {\em ACL\/}.

\end{thebibliography}
\bibliographystyle{acl_natbib}

\appendix
\section{Appendix: Distractor Placement Strategies}
\label{sec:placement}

This section provides a theoretical framework to predict a model's performance
on adversarial test sets when trained on adversarial data generated by a specific
distractor-insertion strategy.

\begin{figure}[h]
    \centering\vspace{-10pt}
    \includegraphics[width=7cm]{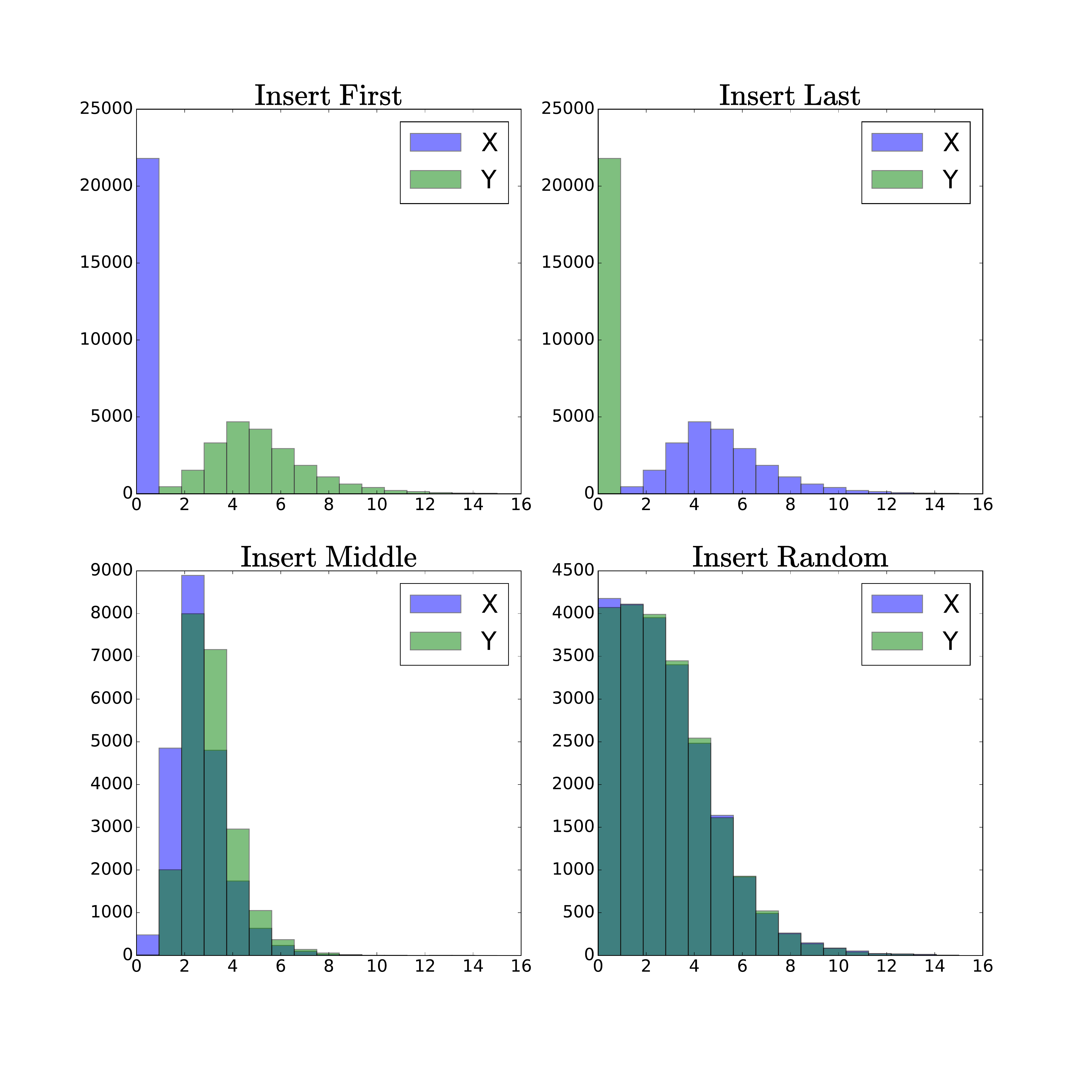}
    \vspace{-30pt}
    \caption{Distributions of $X$ and $Y$ in adversarially augmented SQuAD training
    data under different distractor-insertion strategies.}\label{fig:xy_dist}
    \vspace{-7pt}
\end{figure}

Given a paragraph composed of $n$ sentences (with the distractor inserted)
$P = \{s_1, s_2, \dots, s_n\}$, where $s_i$ is the $i$th sentence
counting from the front. Define random variables $X$ and $Y$ to represent the
location of the distractor counting from the front and back, respectively. The distributions of $X$ and $Y$ are dependent upon the insertion strategy used to add the distractors, several examples of this are displayed in~\figref{fig:xy_dist}.

\begin{figure*}[t]
    \centering
    \begin{minipage}{0.3\textwidth}
        \includegraphics[width=\textwidth]{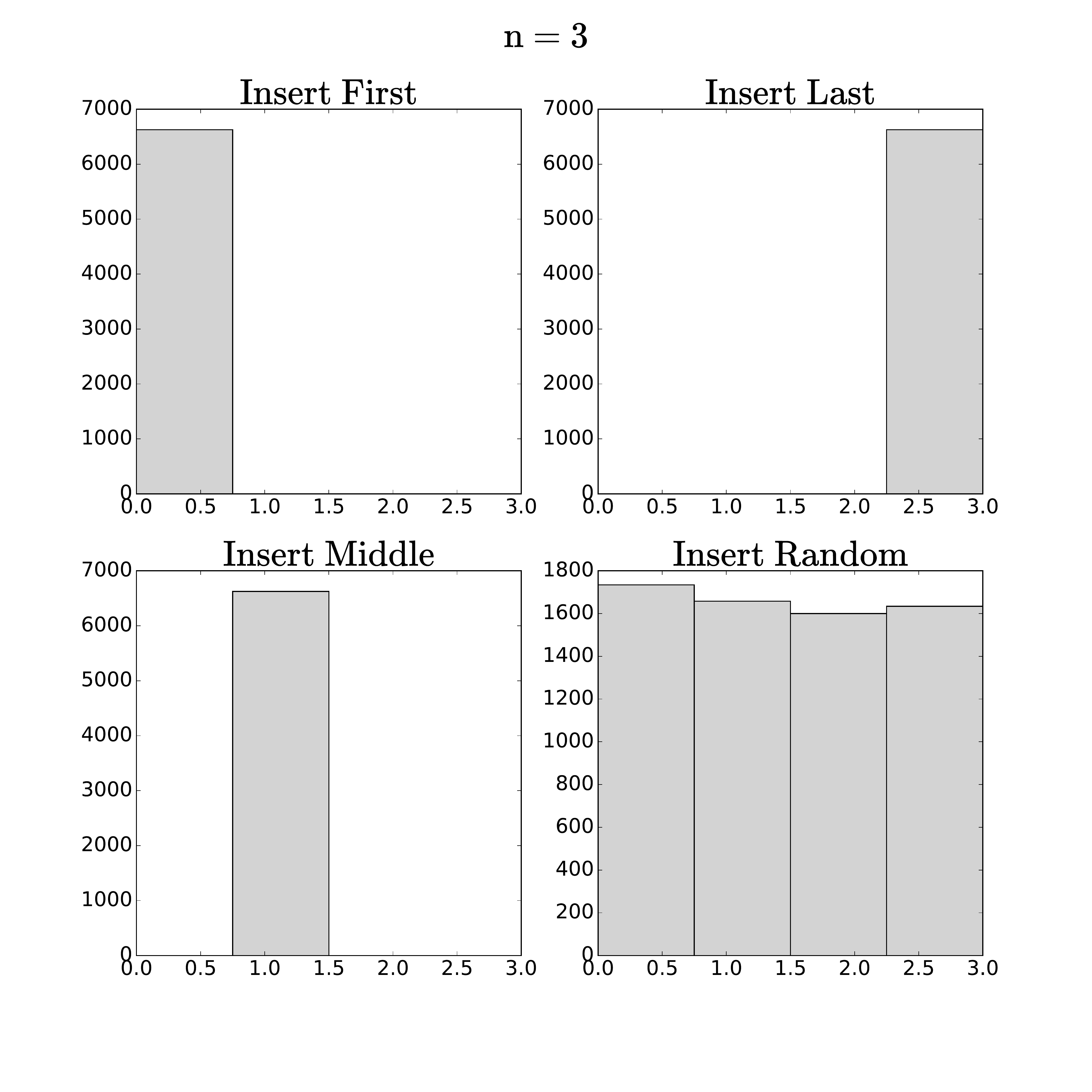}
    \end{minipage}%
    \begin{minipage}{0.3\textwidth}
        \includegraphics[width=\textwidth]{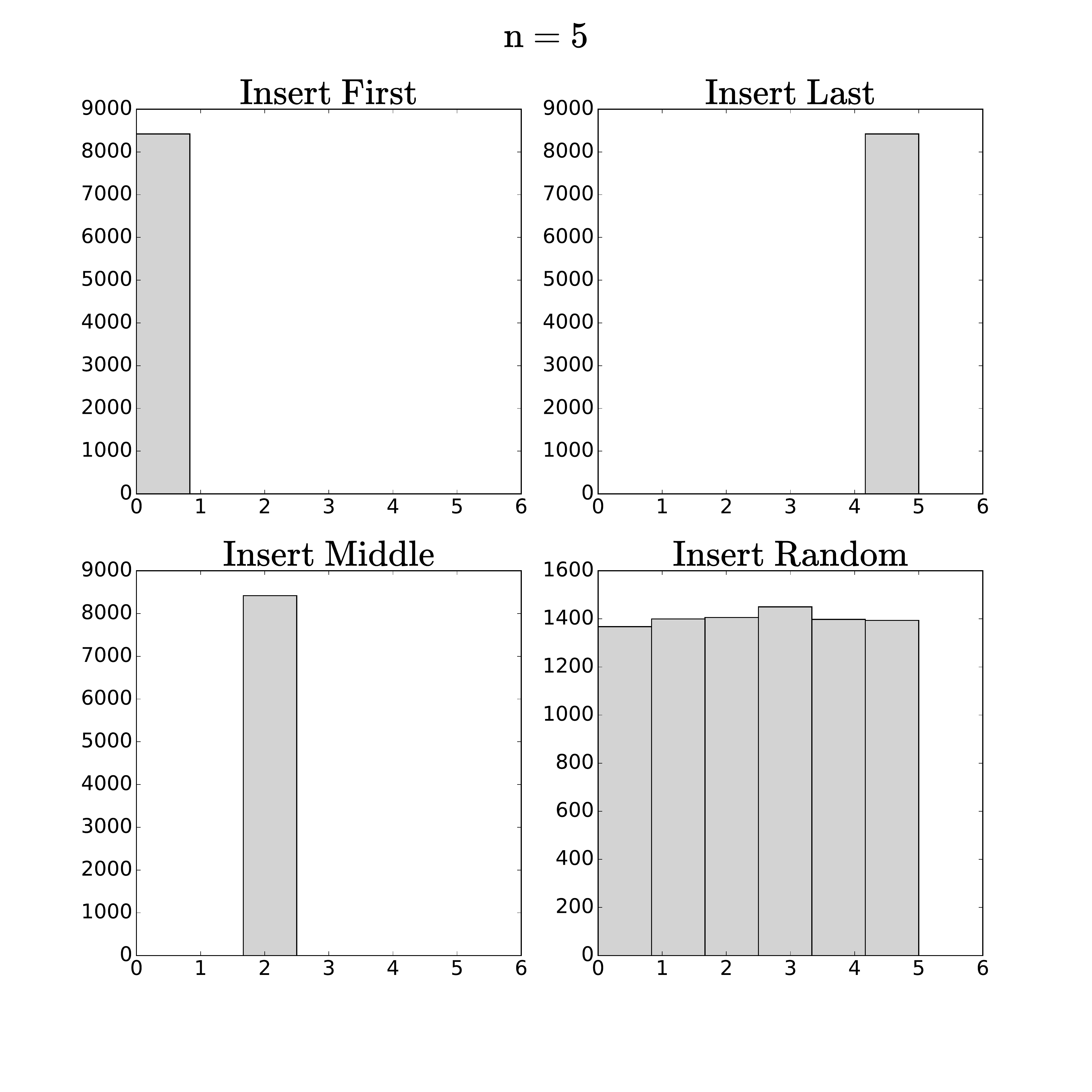}
    \end{minipage}%
    \begin{minipage}{0.3\textwidth}
        \includegraphics[width=\textwidth]{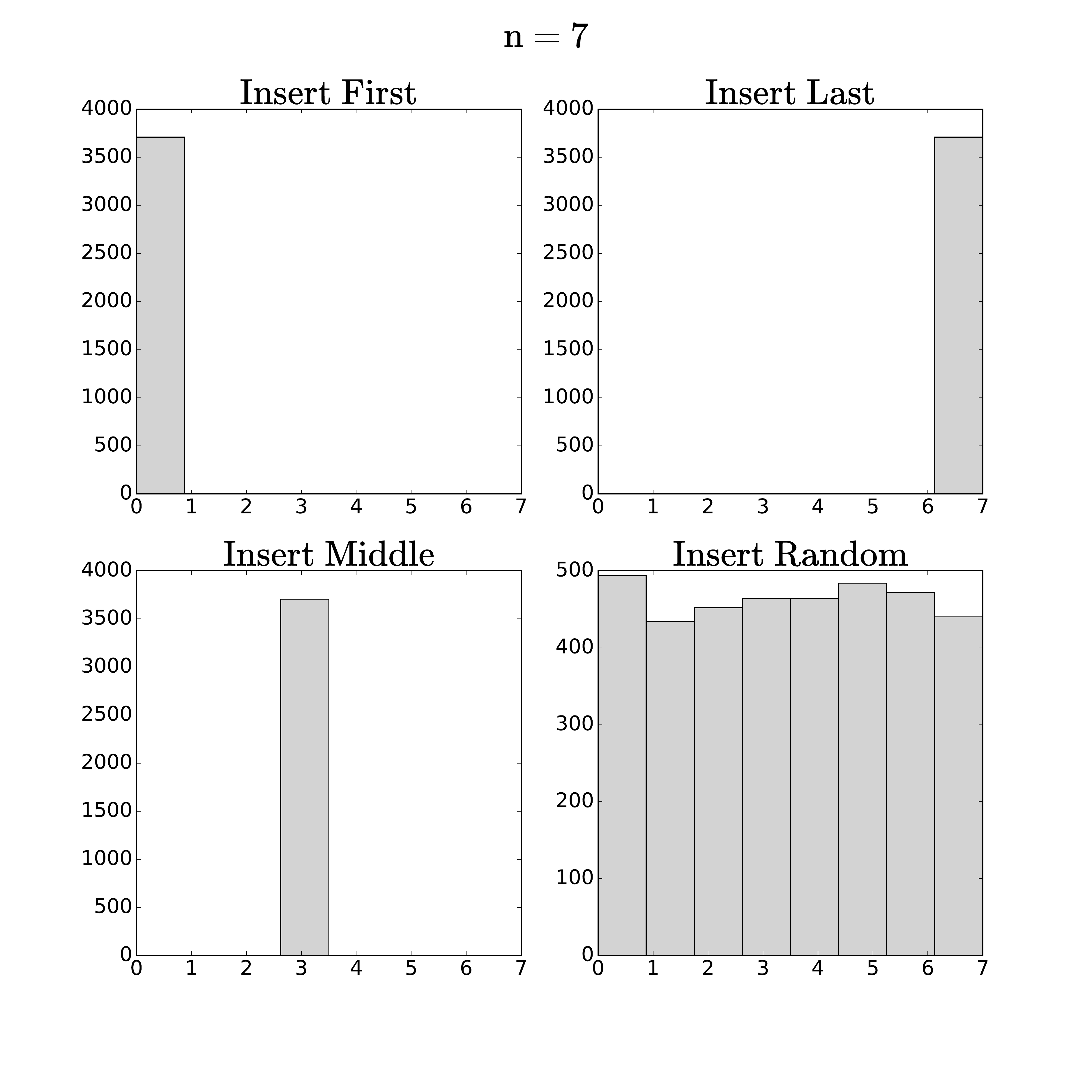}
    \end{minipage}%
    \\
    \begin{minipage}{0.3\textwidth}
        \includegraphics[width=\textwidth]{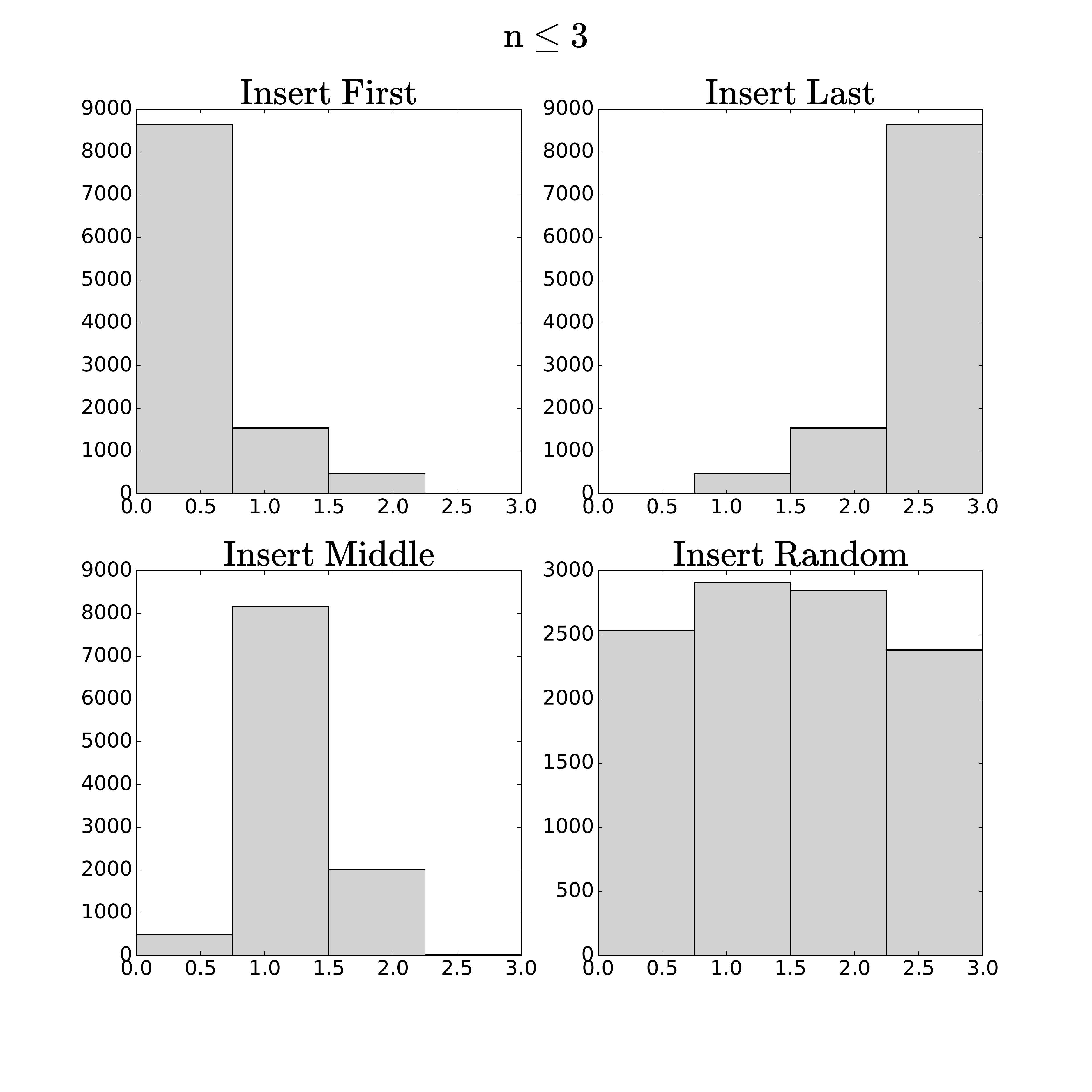}
    \end{minipage}%
    \begin{minipage}{0.3\textwidth}
        \includegraphics[width=\textwidth]{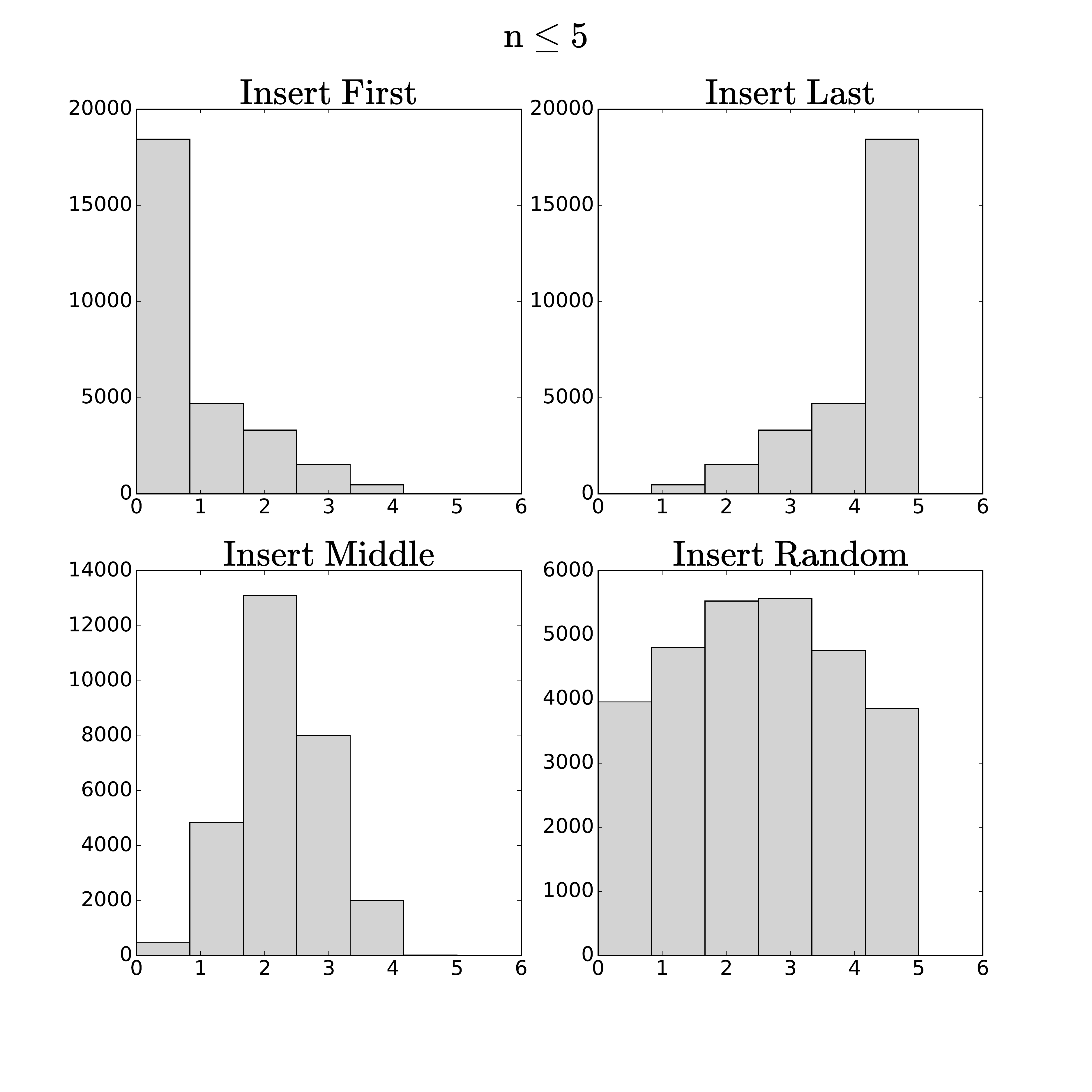}
    \end{minipage}%
    \begin{minipage}{0.3\textwidth}
        \includegraphics[width=\textwidth]{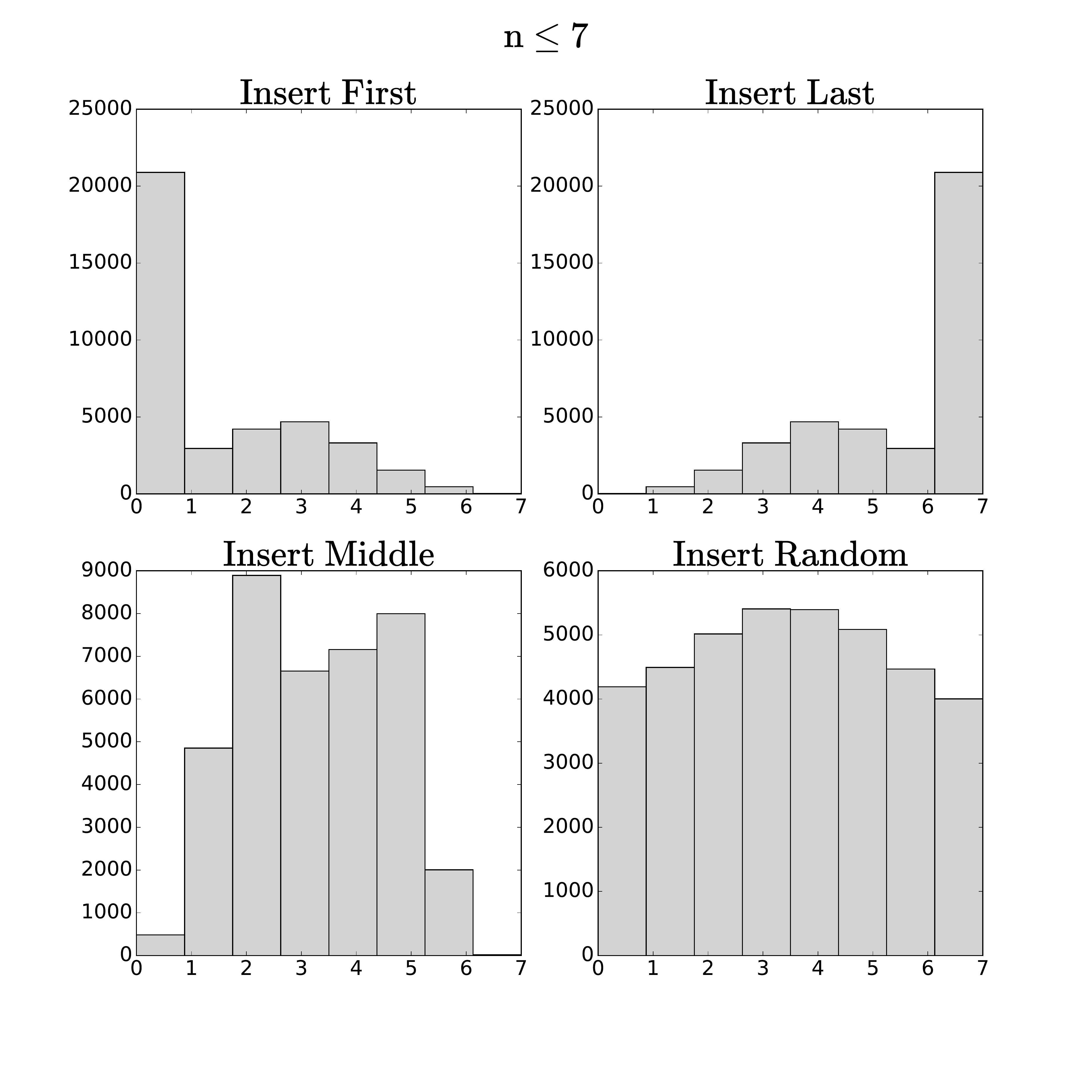}
    \end{minipage}%
    \vspace{-20pt}
    \caption{Learned distribution of $P_{s_a}$ for different $n$.}\label{fig:figure2}
\end{figure*}

\begin{figure}[t]
    \centering
    \includegraphics[width=6cm]{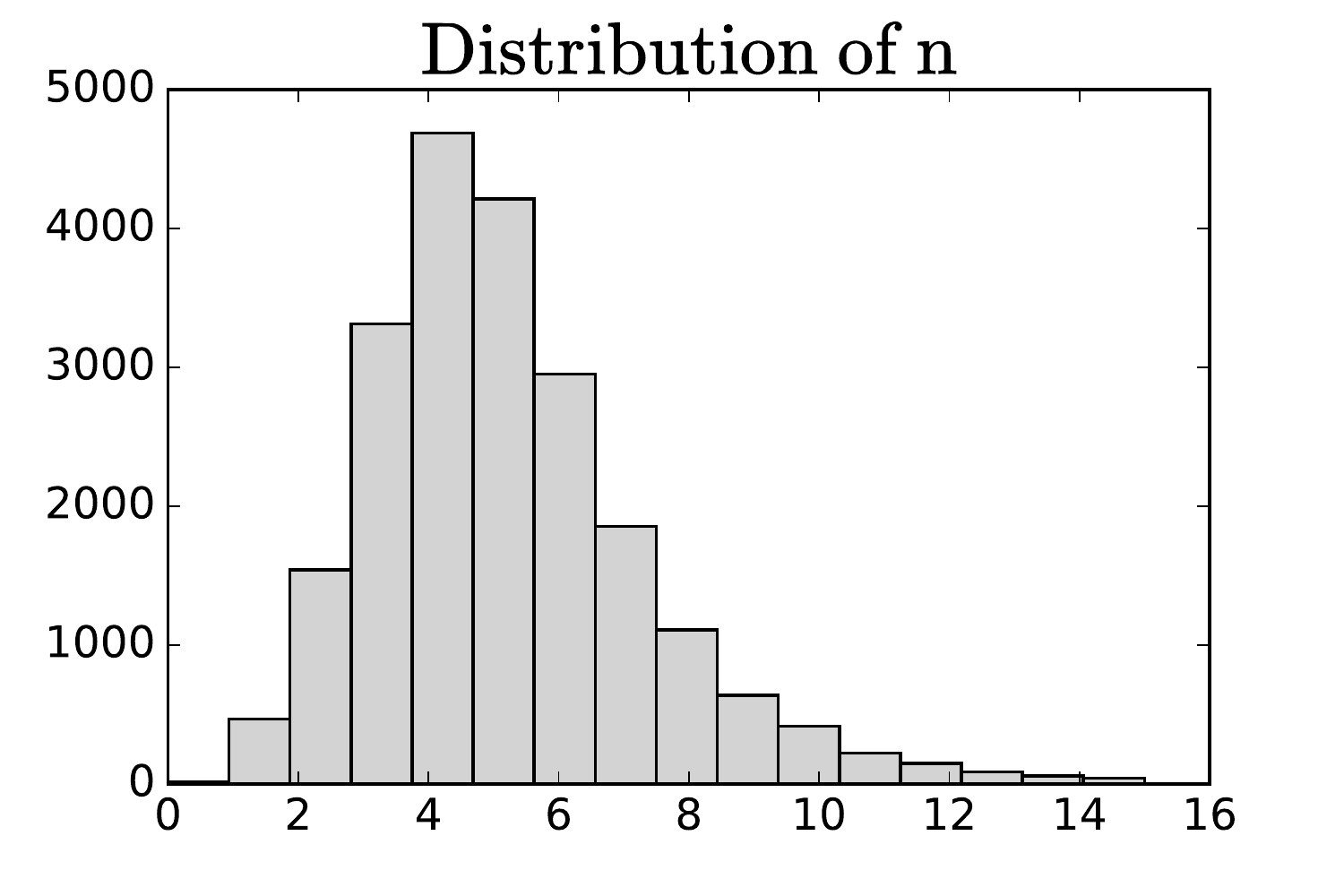}
    \vspace{-15pt}
    \caption{Distribution of length of paragraphs in the SQuAD training
    set.}\label{fig:figure3}
    \vspace{-5pt}
\end{figure}

A bidirectional deep learning model, trained in a supervised setting, should be able to jointly learn $X$ and $Y$. Thus, at test time, when given a paragraph of $n$ sentences, the model can obtain the probability that the sentence $s_a$ is the distractor, $P_{s_a}$, by computing $P(X = a) + P(Y = n - a)$. Ideally, we want the distribution of $P_{s_a}$ to be uniform, as that means the model is not biased towards discarding any sentence as the distractor based on location. The actual distributions of $P_{s_a}$ under different distractor-insertion strategies are displayed in \figref{fig:figure2} for $n = 3$, $5$ and $7$. We pick these $n$ as they are typical lengths of contexts within the SQuAD dataset (the complete distribution of paragraph lengths in the SQuAD training set is shown in \figref{fig:figure3}). We see that under random insertion, the distribution is very close to uniform. Note that if we were to aggregate $n$ and plot $P_{s_a}$ for $n \leq 3$, $5$ and $7$, as shown in \figref{fig:figure2}, the distributions of $P_{s_a}$ created by inserting in the middle and inserting randomly are very similar, but the distribution of inserting in the middle is skewed against the beginnings and ends of the paragraphs. This explains why in our experiment studying the effect of distractor placement strategies (see \tabref{tab:table2}), InsMid's performance was not skewed towards either AddSent or AddSentPrepend, but was worse on both when compared to InsRandom.

\begin{figure}[t]
    \centering
    \includegraphics[clip,width=7cm]{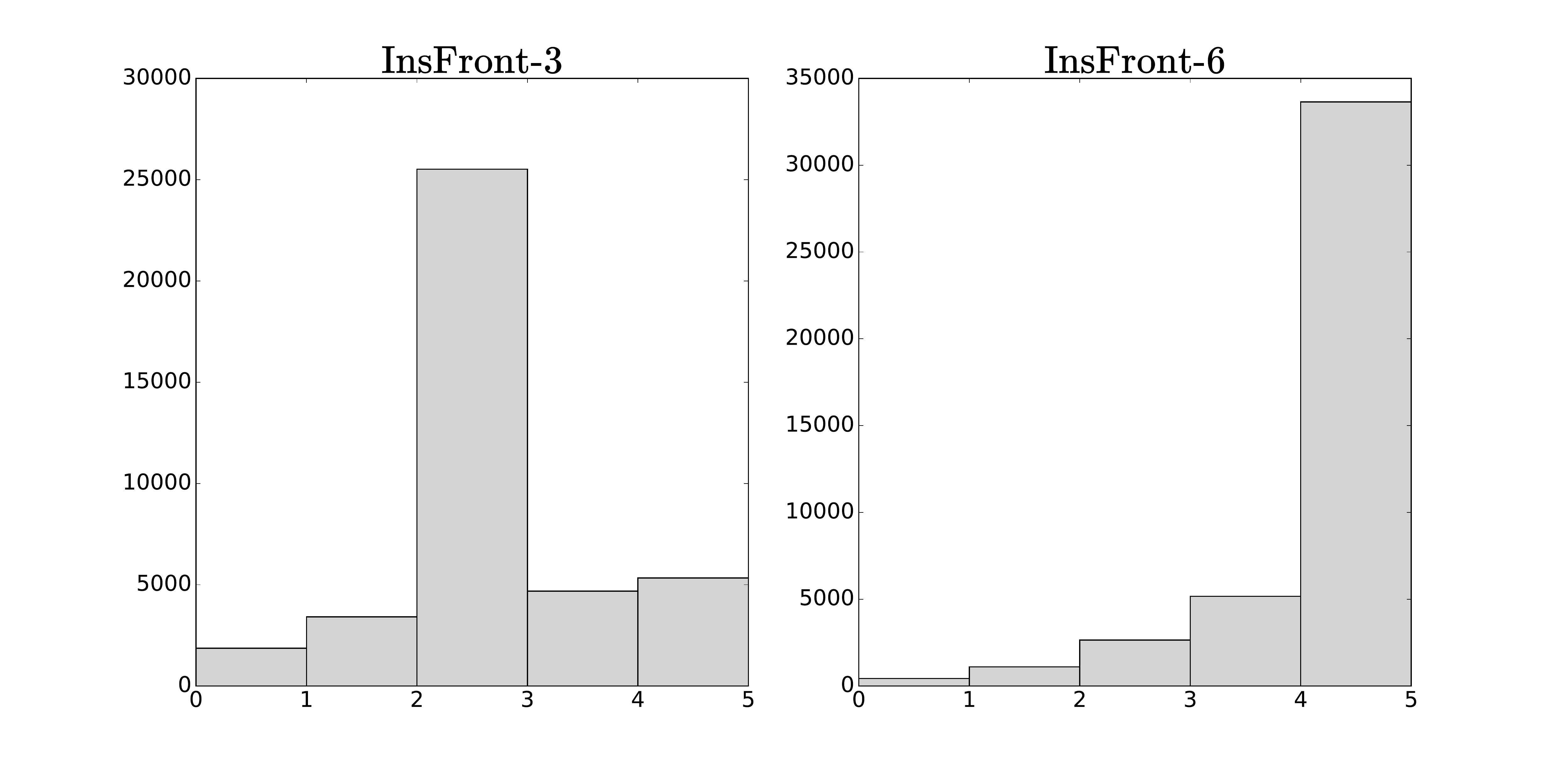}
    \vspace{-15pt}
    \caption{Distributions of $P_{s_a}$ under InsFront-3 and InsFront-6 for $n\leq 5$.}
    \label{fig:figure4}
\end{figure}

This method of calculating the distribution of $P_{s_a}$ allows us to predict the model's 
performance when trained on datasets where the distractors are inserted at specific locations. To test this hypothesis, we created two datasets: InsFront-3 and InsFront-6 where the distractors were inserted as the 3rd and 6th sentence from the beginning and measure the model's performance when trained on these two datasets. The distributions of $P_{s_a}$ for these two datasets are shown in \figref{fig:figure4}, from which we can predict that models trained on InsFront-3 should perform slightly better on adversarial sets where the distractors are appended (as opposed to prepended), whereas those trained on InsFront-6 will perform much better on such adversarial sets. These predictions are confirmed by the results in \tabref{tab:supp_results}.

\begin{table}[t!]
    \centering
    \begin{small}
    \begin{tabular}{c|ccc}
        Training & AddSent & AddSentPrepend & Average \\
        \hline
		InsFront-3 & 75.47 & 72.79 & 74.13 \\
		InsFront-6 & 77.73 & 64.42 & 71.10 \\
    \end{tabular}
    \vspace{-5pt}
    \caption{F1 Performance of the BSAE model trained on datasets with different
    distractor placement strategies.}\label{tab:supp_results}
    \end{small}
\end{table}

\end{document}